\documentclass[a4paper]{article}

\usepackage[utf8]{inputenc}
\usepackage{graphicx}
\usepackage{amsmath}
\usepackage{url}

\begin{document}

\title{Extreme Learning Tree}

\author{Anton Akusok$^1$, Emil Eirola$^1$, Kaj-Mikael Bj\"{o}rk$^2$ \and Amaury Lendasse$^{3,4}$}
\date{%
    $^1$Arcada University of Applied Sciences, Helsinki, Finland\\%
    $^2$Risklab at Arcada UAS, Helsinki, Finland\\
    $^3$Department of Mechanical and Industrial Engineering, \\The University of Iowa, Iowa City, USA
    $^4$The Iowa Informatics Initiative, The University of Iowa, Iowa City, USA
}

\maketitle

\begin{abstract}

The paper proposes a new variant of a decision tree, called an Extreme Learning Tree. It consists of an extremely random tree with non-linear data transformation, and a linear observer that provides predictions based on the leaf index where the data samples fall. The proposed method outperforms linear models on a benchmark dataset, and may be a building block for a future variant of Random Forest.

\end{abstract}

\section{Introduction}

Randomized methods are a recent trend in practical machine learning~\cite{gallicchio_randomized_2017}. They enable the high performance of complex non-linear methods without the high computational cost of their optimization. Current most prominent examples are randomized neural networks, in both feed-forward~\cite{huang_what_2015} and recurrent~\cite{lukosevicius_reservoir_2009} forms. For the latter, the randomized approach provided an efficient training method for the first time, and enabled achieving state-of-the-art performance in multiple areas~\cite{jaeger_harnessing_2004}.

Random forest~\cite{tin_kam_ho_random_1998} is one of the best methods for Big Data processing due to its adaptive nearest neighbour behavior~\cite{lin_random_2006}. The forest predicts an output based only on local data samples. Such an approach works the better the more training data is available, thus making for a perfect supervised method for Big Data. K-nearest neighbors algorithm benefits from more data as the data itself is the model, but Random Forest avoids the quadratic scaling of k-Nearest neighbors in terms of the data samples, that makes it prohibitively slow for large-scale problems.

Decision tree~\cite{breiman_classification_1984} is a building block of Random Forest. A deep decision tree has high variance but low bias. An ensemble of multiple such trees reduces variance, and improves the prediction performance. Additional measures are taken to make the trees in an ensemble as different as possible, including random subsets of features and boosting~\cite{breiman_random_2001}.

The paper proposes a merge between random methods and a decision tree, called an Extreme Learning Tree (ELT). The method builds a tree using expanded data features from an Extreme Learning Machine~\cite{huang_extreme_2012}, by splitting nodes on a random feature at a random point. The result is an Extremely Randomized Tree~\cite{geurts_extremely_2006}. Then a linear observer is added to the leaves of the tree, that learns a linear projection from the leaves to the target outputs. Each tree leaf is represented by its index, in the one-hot encoding format.

\section{Methodology}

Extreme Learning Tree consists of three parts. First, it generates random data features using an Extreme Learning Machine (ELM)~\cite{huang_extreme_2006}. Second, it builds a random tree from these features, similar to Extremely Randomized Trees~\cite{geurts_extremely_2006}. Each data sample is then represented by the index of its leaf from the tree, in one-hot encoding. Third, a linear regression is learned from the dataset in that one-hot encoding to the target outputs.

ELT follows the random methods paradigm as it has an untrained random part (the tree), and a learned linear observer (a linear regression model from leaves of the tree to the target outputs).

An ELT tree has two hyper parameters: the minimum node size, and the maximum thee depth. A node data is split by a random feature using a random split point. Split points that generates nodes under the minimum size are rejected. Nodes that reach the maximum depth or under twice the minimum size become leafs. Node splitting continues until there are non-leaf terminal nodes.

\section{Experimental results}

The Extreme Learning Tree is tested on well-known Iris flower dataset~\cite{fisher_use_1936}, in comparison with a Decision Tree, an L2 regularized ELM~\cite{miche_tropelm_2011}, and Ridge regression. Decision Tree implementation is from the Scikit-Learn library\footnote{\url{http://scikit-learn.org/stable/auto_examples/tree/plot_iris.html}}.

\begin{figure}[ht]
    \centering
    \includegraphics[width=0.6\textwidth]{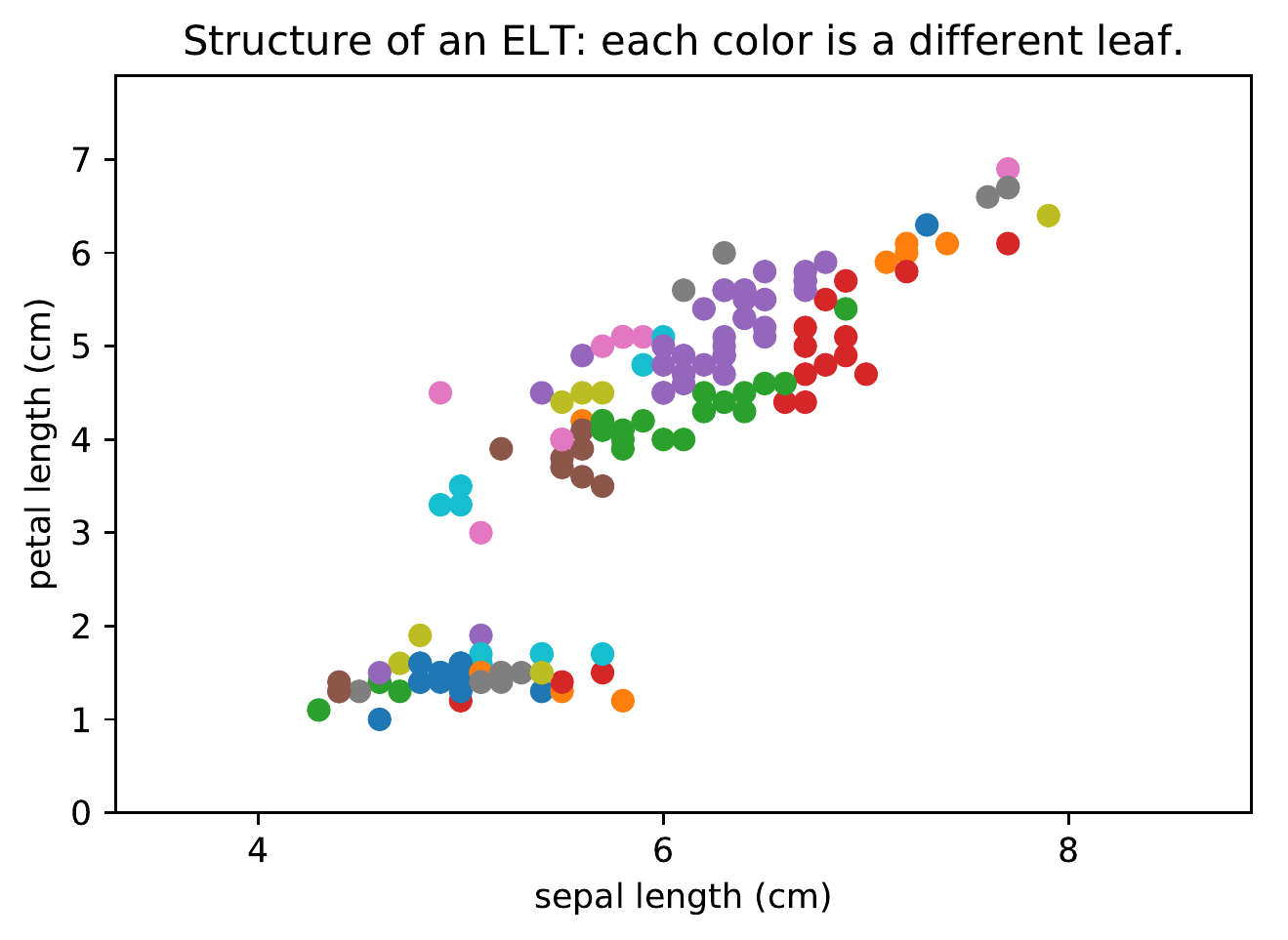}
    \caption{Leaf structure of an ELT, each color represents a different leaf. The random tree works as an approximated nearest neighbour method, joining together similar data samples.}
    \label{fig:structure}
\end{figure}

The random tree in the ELT method splits data samples into groups of similar ones. The resulting structure in the original data space is shown on Figure~\ref{fig:structure}. The tree works as a adaptive nearest neighbour, combining together similar samples. Then the target variable information from these samples is used by a linear observer to make predictions.

A formal performance comparison is done on Iris dataset. The data is randomly split into 70\% training and 30\% test sets, and the test accuracy is calculated for all the methods. The whole experiment is repeated 100 times. Mean accuracy and its standard deviation are presented in Table~\ref{tab:1}.
\begin{table}[ht]
    \centering
    \caption{Average accuracy and its standard deviation on a test subset of Iris dataset.}
    \label{tab:1}
    \begin{tabular}{c|c}
        Method & Accuracy $\pm$ std, \% \\
        \hline
        Ridge regression & $82.7 \pm 5.1$ \\
        Extreme Learning Tree & $87.2 \pm 6.1$ \\
        ELM & $90.9 \pm 5.0$ \\
        Decision Tree & $94.1 \pm 3.2$
    \end{tabular}
\end{table}

In this experiment, an Extreme Learning Tree performs under ELM and Decision Tree methods. However, it outperforms a linear model (in the form of Ridge regression) by a significant margin. Outperforming a linear model is an achievement for a single ELT, as it represents each data sample by a single number -- an index of its leaf in the tree.

\begin{figure}
    \centering
    \includegraphics[width=\textwidth]{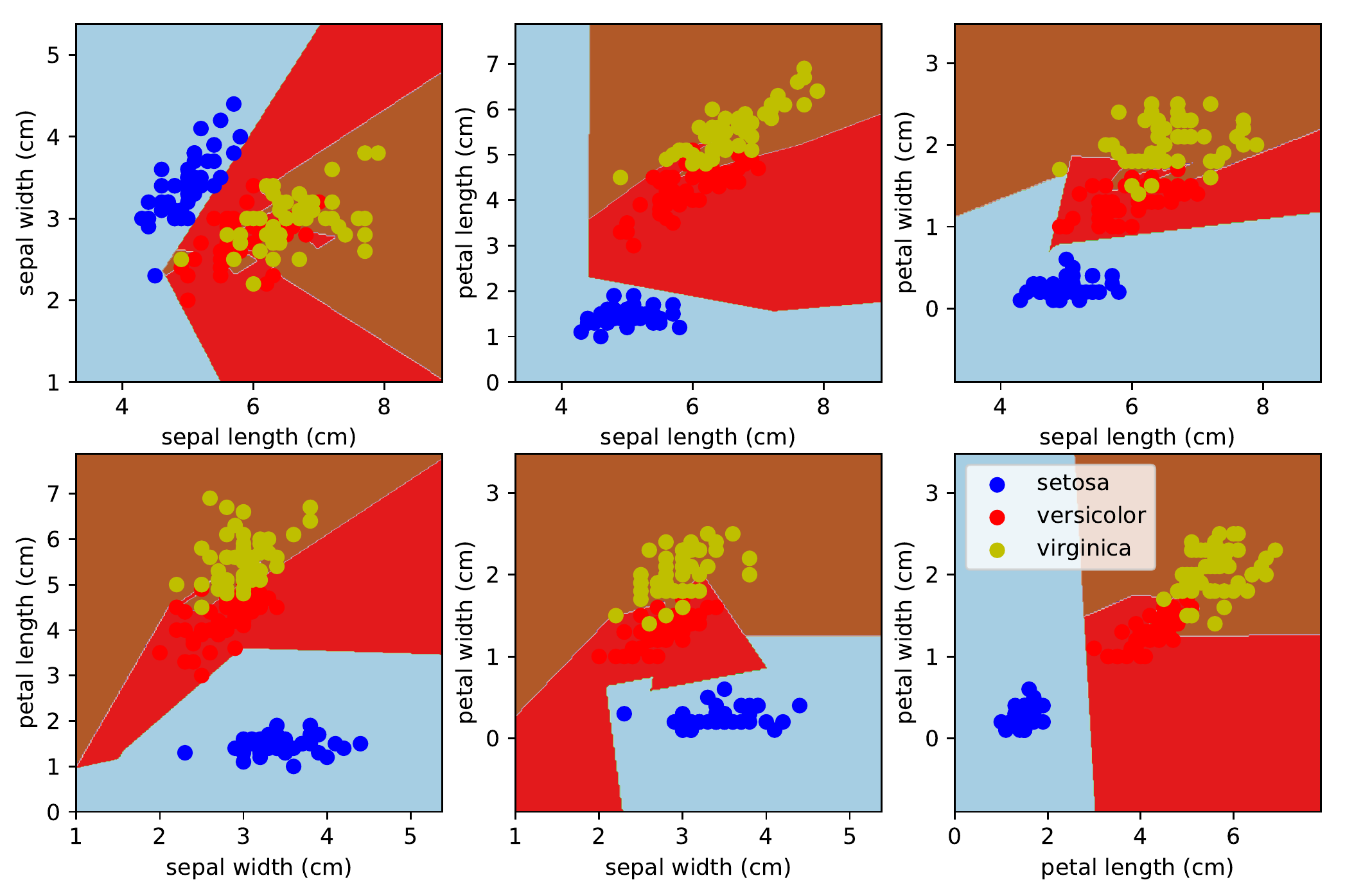}
    \caption{Decision surface of an ELT on Iris dataset, using different pairs of features. Different colors correspond to the three different classes of Iris flowers.}
    \label{fig:elt}
    
    \vspace*{\floatsep}
    
    \centering
    \includegraphics[width=\textwidth]{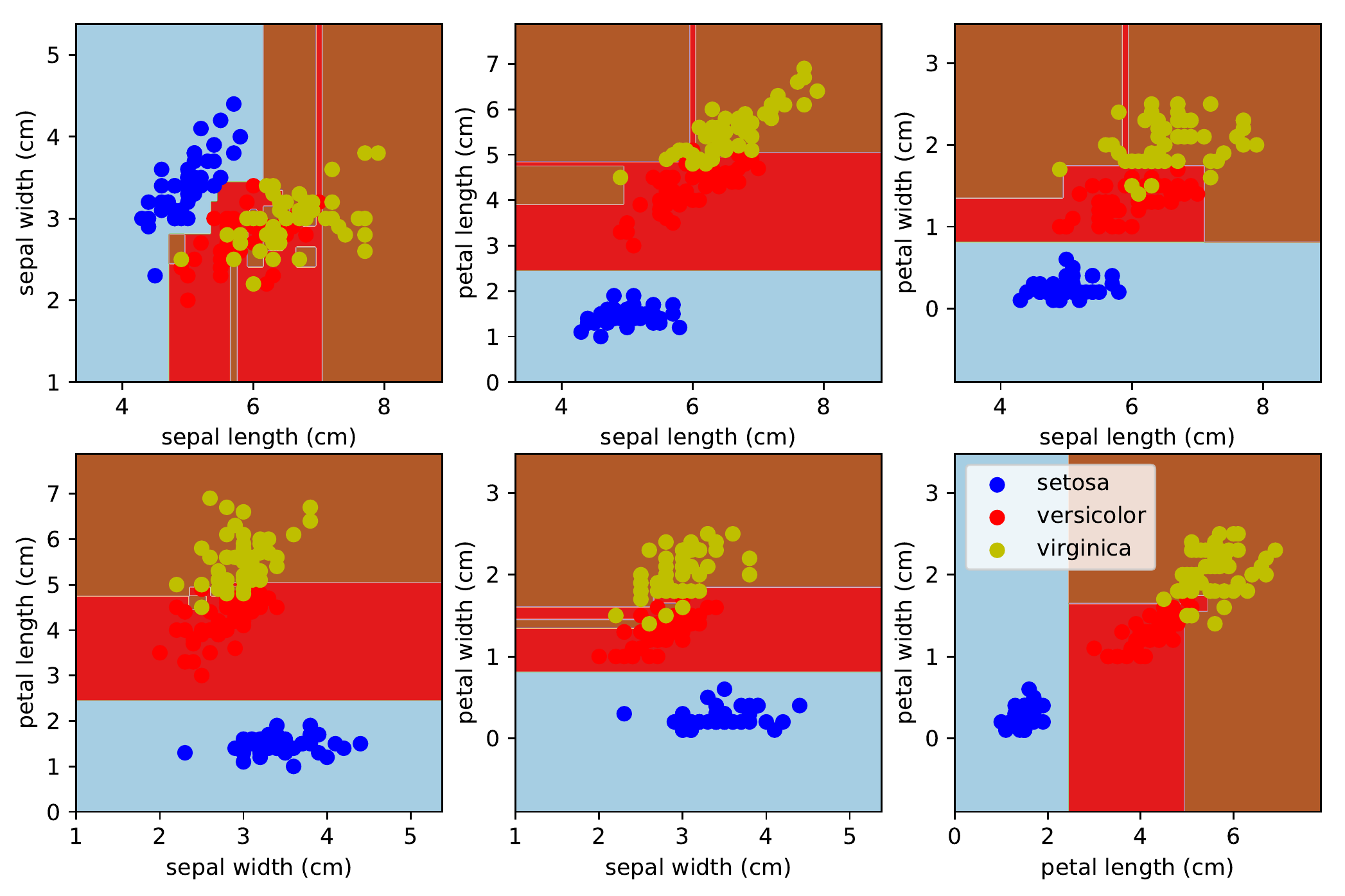}
    \caption{Decision surface of a Decision Tree on Iris dataset, using different pairs of features. Note that all decision boundaries are parallel to axes.}
    \label{fig:dt}
\end{figure}

Decision surface of ELT is visualized on Figure~\ref{fig:elt}. The boundaries between classes have complex shape, but the classes are unbroken. Class boundaries of the original Decision Tree (shown on Figure~\ref{fig:dt}) break into each other creating false predictions. They are always parallel to an axis, while ELT learns class boundaries of an arbitrary shape.

\section{Conclusions}

The paper proposes a new version of decision tree, that follows the random methods paradigm. It consists of an untrained random non-linear tree, and a learned linear observer. The method provides decision boundaries of a complex shape and with less noise than an original decision tree. It outperforms a purely linear model in accuracy despite representing the data samples only by a corresponding tree leaf index.

Future works will examine an application of Extreme Learning Tree to an ensemble method similar to Random Forest.

\bibliographystyle{plain}
\bibliography{elt}

\begin{thebibliography}{10}

\bibitem{gallicchio_randomized_2017}
Gallicchio, C., Martin-Guerrero, J.D., Micheli, A., Soria-Olivas, E.:
\newblock Randomized machine learning approaches: {{Recent}} developments and
  challenges.
\newblock In: {{ESANN}} 2017 Proceedings, {{European Symposium}} on
  {{Artificial Neural Networks}}, {{Computational Intelligence}} and {{Machine
  Learning}}.
\newblock {d-side publi.}, Bruges, Belgium (26-28 April 2017)  77--86

\bibitem{huang_what_2015}
Huang, G.B.:
\newblock What are {{Extreme Learning Machines}}? {{Filling}} the {{Gap Between
  Frank Rosenblatt}}'s {{Dream}} and {{John}} von {{Neumann}}'s {{Puzzle}}.
\newblock Cognitive Computation \textbf{7}(3) (2015)  263--278

\bibitem{lukosevicius_reservoir_2009}
Luko{\v s}evi{\v c}ius, M., Jaeger, H.:
\newblock Reservoir computing approaches to recurrent neural network training.
\newblock Computer Science Review \textbf{3}(3) (August 2009)  127--149

\bibitem{jaeger_harnessing_2004}
Jaeger, H., Haas, H.:
\newblock Harnessing {{Nonlinearity}}: {{Predicting Chaotic Systems}} and
  {{Saving Energy}} in {{Wireless Communication}}.
\newblock Science \textbf{304}(5667) (April 2004) ~78

\bibitem{tin_kam_ho_random_1998}
{Tin Kam Ho}:
\newblock The random subspace method for constructing decision forests.
\newblock IEEE Transactions on Pattern Analysis and Machine Intelligence
  \textbf{20}(8) (August 1998)  832--844

\bibitem{lin_random_2006}
Lin, Y., Jeon, Y.:
\newblock Random {{Forests}} and {{Adaptive Nearest Neighbors}}.
\newblock Journal of the American Statistical Association \textbf{101}(474)
  (June 2006)  578--590

\bibitem{breiman_classification_1984}
Breiman, L., Friedman, J., Stone, C.J., Olshen, R.A.:
\newblock Classification and Regression Trees.
\newblock {CRC press} (1984)

\bibitem{breiman_random_2001}
Breiman, L.:
\newblock Random {{Forests}}.
\newblock Machine Learning \textbf{45}(1) (2001)  5--32

\bibitem{huang_extreme_2012}
Huang, G.B., Zhou, H., Ding, X., Zhang, R.:
\newblock Extreme learning machine for regression and multiclass
  classification.
\newblock Systems, Man, and Cybernetics, Part B: Cybernetics, IEEE Transactions
  on \textbf{42}(2) (April 2012)  513--529

\bibitem{geurts_extremely_2006}
Geurts, P., Ernst, D., Wehenkel, L.:
\newblock Extremely randomized trees.
\newblock Machine Learning \textbf{63}(1) (2006)  3--42

\bibitem{huang_extreme_2006}
Huang, G.B., Zhu, Q.Y., Siew, C.K.:
\newblock Extreme learning machine: {{Theory}} and applications.
\newblock Neural Networks Selected Papers from the 7th Brazilian Symposium on
  Neural Networks (SBRN '04)7th Brazilian Symposium on Neural Networks
  \textbf{70}(1\textendash{}3) (December 2006)  489--501

\bibitem{fisher_use_1936}
FISHER, R.A.:
\newblock {{THE USE OF MULTIPLE MEASUREMENTS IN TAXONOMIC PROBLEMS}}.
\newblock Annals of Eugenics \textbf{7}(2) (1936)  179--188

\bibitem{miche_tropelm_2011}
Miche, Y., {van Heeswijk}, M., Bas, P., Simula, O., Lendasse, A.:
\newblock {{TROP}}-{{ELM}}: {{A}} double-regularized {{ELM}} using {{LARS}} and
  {{Tikhonov}} regularization.
\newblock Advances in Extreme Learning Machine: Theory and Applications
  Biological Inspired Systems. Computational and Ambient Intelligence Selected
  papers of the 10th International Work-Conference on Artificial Neural
  Networks (IWANN2009) \textbf{74}(16) (September 2011)  2413--2421

\end{thebibliography}

\end{document}